\documentclass[conference]{IEEEtran}
\IEEEoverridecommandlockouts
\usepackage{cite}
\usepackage{amsmath,amssymb,amsfonts}
\usepackage{algorithmic}
\usepackage{algorithm}
\usepackage{graphicx}
\usepackage{textcomp}
\usepackage{hyperref}
\usepackage{xcolor}
\usepackage{pythontex} 
\usepackage{booktabs}
\def\BibTeX{{\rm B\kern-.05em{\sc i\kern-.025em b}\kern-.08em
    T\kern-.1667em\lower.7ex\hbox{E}\kern-.125emX}}
\begin{document}

\IEEEoverridecommandlockouts
% \IEEEpubid{\makebox[\columnwidth]{978-1-6654-8045-1/22/\$31.00 \copyright 2022 IEEE \hfill}
% \hspace{\columnsep}\makebox[\columnwidth]{ }}

\title{TabIQA: Table Questions Answering on \\ Business Document Images
%*\\
}
\author{
\IEEEauthorblockN{Phuc Nguyen\textsuperscript{\textsection}, Nam Tuan Ly\textsuperscript{\textsection}, Hideaki Takeda, Atsuhiro Takasu}
\IEEEauthorblockA{\textit{National Institute of Informatics, Japan}\\
\{phucnt, namly, takeda, takasu\}@nii.ac.jp}
}

\maketitle
\begingroup
\renewcommand\thefootnote{\textsection}
\footnotetext{Equal contribution}
\endgroup
\begin{abstract}
Table answering questions from business documents has many challenges that require understanding tabular structures, cross-document referencing, and additional numeric computations beyond simple search queries. This paper introduces a novel pipeline, named TabIQA, to answer questions about business document images. TabIQA combines state-of-the-art deep learning techniques 1) to extract table content and structural information from images and 2) to answer various questions related to numerical data, text-based information, and complex queries from structured tables. The evaluation results on VQAonBD 2023 dataset demonstrate the effectiveness of TabIQA in achieving promising performance in answering table-related questions. The TabIQA repository is available at \url{https://github.com/phucty/itabqa}. 
\end{abstract}

\begin{IEEEkeywords}
Visual Question Answering, Table Question Answering, Business Documents
\end{IEEEkeywords}

\section{Introduction} \label{sec:intro}
% Background:
Businesses generate and process vast amounts of information, and extracting valuable insights from this data is crucial for making informed decisions. Business documents, such as financial reports, invoices, and contracts, often contain valuable information in tabular form. However, answering questions based on these document images can be challenging due to their complex structures, cross-referencing, and numerical computations beyond simple search queries.

% Motivation
Traditional information retrieval methods, such as keyword search and regular expressions, are only sometimes effective in retrieving information from tables in business documents. Therefore, there is a growing need for automated approaches that can accurately extract relevant information from tables and answer various questions. Such approaches can save businesses time and effort and improve the extracted information's accuracy and reliability.

\begin{figure}[htbp]
\centering
\includegraphics{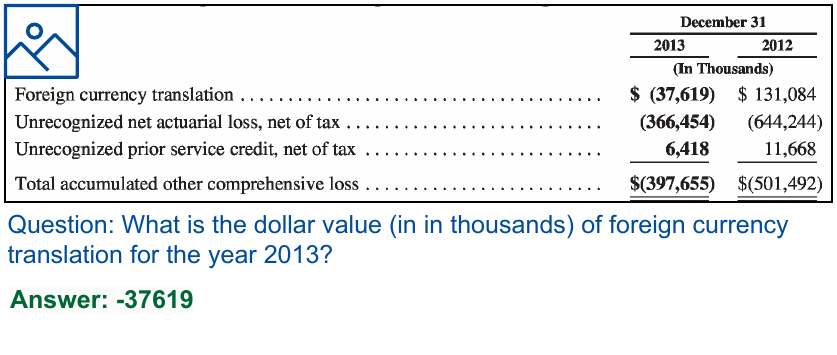}
\caption{Illustration of question answering from the business document image of \text{val\_table\_image\_7517\_\_GPC\_\_2013\_\_page\_55\_split\_0} from VQAonBD 2023 dataset}
\label{fig:fig1}
\end{figure}

% Objectives and methods
The main objective of this paper is to introduce TabIQA, a novel pipeline for answering questions on business document images. Figure \ref{fig:fig1} illustrates the question-answering task from the business document image in VQAonBD 2023 dataset. Given a business document image and a question about the image: ``What is the dollar value (in thousands) of foreign currency translation for the year 2013?" the output answer is  ``-37619". TabIQA utilizes the table recognition module to extract table structure information and the text content of each table cell and convert them into HTML format. Subsequently, the high-level table structure is extracted to identify the headers, data cells, and hierarchical structure with the post-structure extraction module. Once the table is structured, it is converted to a dataframe format for further processing. The question-answering module processes the input question and the table dataframe with an encoder and generates the final answer from a decoder.  

% Contributions
Overall, this study makes the following contributions:
\begin{itemize}
    \item Introducing TabIQA, a novel pipeline for answering questions about business document images: TabIQA is a comprehensive pipeline combining state-of-the-art deep learning techniques to extract relevant information from tables and answer various questions related to numerical data, text-based information, and complex queries. 
    \item Providing a publicly available repository: We have made the TabIQA repository publicly available to encourage the reproducibility of our results and enable other researchers to use and build upon our work.
    \item Demonstrating the effectiveness of TabIQA on the VQAonBD 2023 dataset: The evaluation results on VQAonBD 2023 dataset\footnote{VQAonBD 2023: \url{https://ilocr.iiit.ac.in/vqabd/dataset.html}} demonstrate the effectiveness of TabIQA in achieving promising performance in answering table-related questions.
\end{itemize}

The rest of the paper is structured as follows. Section \ref{sec2:framework} summarizes related work on question answering on business document images. We introduce the TabIQA method in Section \ref{sec1:approach}. Section \ref{sec1:experiments} presents the experimental settings and results. Finally, in Section \ref{sec1:conclusion}, we present conclusions and discuss future directions of the task table question answering on business document images. 

\section{Related Work} \label{sec1:related_work}
Image-based table recognition is one of the important parts of the document understanding system as well as the table questions answering system, which aims to recognize the table structure information and the text content of each table cell from an input image and represent them in a machine-readable format (HTML or CSV). Most of the previous works \cite{QiaoICDAR2021, Ye2021PingAnVCGroupsSF, NassarCVPR2022, ZhangPR2022} of table recognition focused on two-step approaches that divide the problem into two sub-problems: table structure recognition and table cell content recognition, and then attempt to solve each sub-problem independently by two separate systems. In recent years, due to the advantages of deep learning and the availability of large-scale table image datasets, some works \cite{Deng2019, Zhong2020, multitasktabnet, wstabnet} try to focus on end-to-end approaches which solve the table recognition problem using one end-to-end system. Ly et al. \cite{multitasktabnet} formulated the problem of table recognition as a multi-task learning problem. They proposed an end-to-end multi-task learning model for image-based table recognition, which consists of three separate decoders for three sub-tasks of table recognition: table structure recognition, cell detection, and cell-content recognition. The proposed model achieves state-of-the-art accuracies on PubTabNet and FinTabNet datasets. Ly et al. \cite{wstabnet} also proposed an end-to-end weakly supervised learning model named WSTabNet, which requires only table images and their annotations of table HTML code for training the model. WSTabNet achieves competitive accuracies compared to the fully supervised and two-step learning methods.

Information Retrieval from Business Documents Business documents such as invoices, receipts, and financial statements contain valuable information critical for decision-making and analysis. Traditional information retrieval methods from business documents rely on manual data entry or simple search queries, which can be time-consuming and error-prone. Recent advances in deep learning and natural language processing have enabled the development of automated systems that can extract information from business documents with high accuracy and efficiency.

Deep Learning Techniques for Table Extraction and Question Answering Table extraction and question answering are critical tasks in automated information retrieval from business documents. Deep learning techniques have shown great promise in addressing these challenges. Various approaches have been proposed for table extraction, including region-based, cell-based, and structure-based methods. For question answering, neural network-based models such as transformer-based models have achieved state-of-the-art results on various datasets.

While these studies have achieved promising results in automated information retrieval from business documents, there is still room for improvement in accuracy, efficiency, and scalability. This paper proposes a novel pipeline, TabIQA, for answering questions about business document tables that leverage state-of-the-art deep learning techniques for improved performance.

\section{Approach} \label{sec1:approach}
This section describes TabIQA's overall framework in Section \ref{sec2:framework}. The details of the table recognition module, post-structure extraction module, and question-answering module are described in Section \ref{sec2:table_recognition}, \ref{sec2:structure_extraction}, and Section \ref{sec2:qa} respectively. 

\subsection{Framework} \label{sec2:framework}
\begin{figure}[htbp]
\centering
\includegraphics{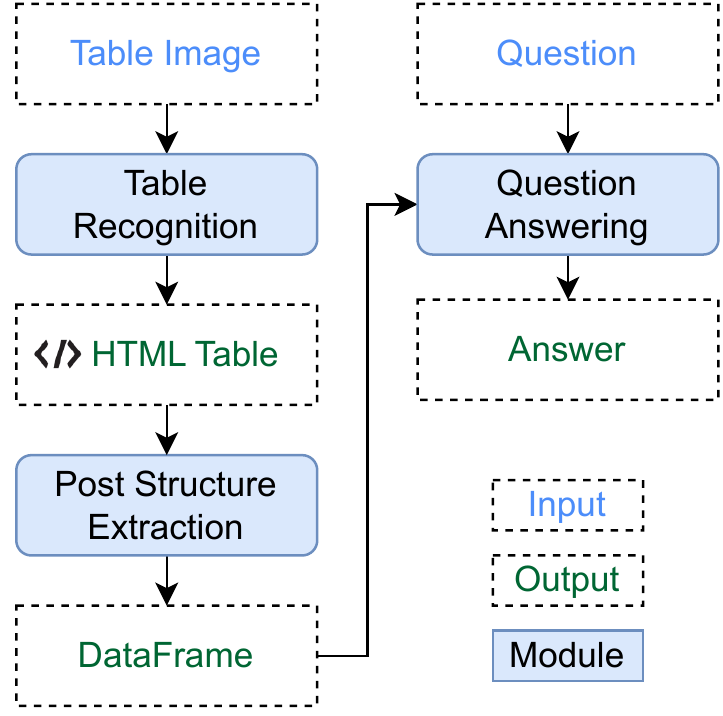}
\caption{Framework}
\label{fig:framework}
\end{figure}

The overall framework of TabIQA, a system designed for question-answering using table images in business documents, is illustrated in Fig. \ref{fig:framework}. TabIQA utilizes a table recognition algorithm that extracts the table's structure and textual content of each cell and then converts them into HTML format. The system subsequently analyzes the HTML table to identify headers, data cells, and hierarchical structure and transforms it into a dataframe for further processing. Finally, the question-answering module processes the input question and the table dataframe with an encoder and generates the final answer through a decoder. 

\subsection{Table Recognition} \label{sec2:table_recognition}
This module aims to predict the table structure information and the text content of each table cell from a table image and represent them in a machine-readable format (HTML). This module consists of one shared encoder, one shared decoder, and three separate decoders for three sub-tasks of table recognition: table structure recognition, cell detection, and cell-content recognition.

First, we trained this model on the training set of VQAonBD 2023 and validated it on the validation set for model selection and choosing the hyperparameters. Finally, we used training and validation sets to train the final table recognition module.

\subsection{Post Structure Extraction} \label{sec2:structure_extraction}
The table post-structure extraction module plays a crucial role in the TabIQA system. The module's primary function is to predict table headers and extract the hierarchical rows from the HTML table.
\subsubsection{Header Prediction}
To predict table headers, the module uses a set of heuristics based on the characteristics of the input table. Specifically, the headers are identified as one of the first table rows that satisfy one or more of the following conditions:
\begin{itemize}
    \item Column spans: The header row contains cells that span multiple columns. 
    \item Nan cells: The header row contains cells with missing values.
    \item Duplicate value cells: The header row contains cells with identical values in the same row.
    \item If no headers are found using these heuristics, the first row is treated as the table header.
\end{itemize}
All the remaining rows are then classified as data rows.

\subsubsection{Hierarchical Row Prediction}
This module predicts the hierarchical information from the table HTML and then concatenates the value of each hierarchical cell to the lower-level cells in the same column. In this work, we propose two hierarchical row prediction algorithms: the first one is based on the predicted table HTML and the second one is based on both the predicted table HTML and the table cell bounding boxes. The two hierarchical row prediction algorithms are defined in the algorithm 1 and 2, respectively:

\begin{algorithm}
\caption{Hierarchical row prediction algorithm}
\begin{algorithmic}
\renewcommand{\algorithmicrequire}{\textbf{Input:}}
\renewcommand{\algorithmicensure}{\textbf{Output:}}
\REQUIRE R (a list of rows of table cells)
\ENSURE  R' (a list of rows of table cells with hierarchical information)
\STATE $hierarchical\_cell \gets null$
\FOR{r in R}
    \IF{r[0] is a colspan cell OR r[1] is an empty cell}
        \STATE $hierarchical\_cell \gets r[0]$
    \ELSIF{r[0] is a rowspan cell OR r[0] is an empty cell}
        \STATE $hierarchical\_cell \gets null$
    \ELSIF{hierarchical\_cell is not null}
        \STATE $r[0].text \gets r[0].text + hierarchical\_cell.text$
    \ENDIF
\ENDFOR
\RETURN $R$
\end{algorithmic}
\end{algorithm}

\begin{algorithm}
\caption{Hierarchical row prediction algorithm}
\begin{algorithmic}
\renewcommand{\algorithmicrequire}{\textbf{Input:}}
\renewcommand{\algorithmicensure}{\textbf{Output:}}
\REQUIRE R (a list of rows of table cells with their bounding boxes)
\ENSURE  R' (a list of rows of table cells with hierarchical information)
\STATE $hierarchical\_cell \gets null$
\STATE $different\_bbox\_flag \gets False$
\FOR{r in R}
    \IF{r[0] is a colspan cell OR r[1] is an empty cell}
        \STATE $hierarchical\_cell \gets r[0]$
    \ELSIF{r[0] is a rowspan cell OR r[0] is an empty cell}
        \STATE $hierarchical\_cell \gets null$
    \ELSIF{$different\_bbox\_flag$ AND r[0].left\_bbox - $hierarchical\_cell$.left\_bbox < threshold}
        \STATE $hierarchical\_cell \gets null$
    \ELSIF{hierarchical\_cell is not null}
        \STATE $r[0].text \gets r[0].text + hierarchical\_cell.text$
    \ENDIF
\ENDFOR
\RETURN $R$
\end{algorithmic}
\end{algorithm}

\subsection{Question Answering} \label{sec2:qa}
We adopt the state-of-the-art table-based QA setting OmniTab \cite{omnitab}, based on TAPEX \cite{tapex} pre-training setting. It feeds the concatenating token sequences of natural language questions and linearized table dataframes into the bidirectional encoder. The table dataframes are linearized in the order of top-to-bottom and left-to-right. The final answers are generated with an autoregressive decoder.

We create a new fine-tuning training set from the training set of VQAonBD 2023 and the results of table recognization and post-structure extraction modules. Each sample consists of a table dataframe and a natural language question, the equivalent ground truth answer. We fine-tuned the OmniTab large pre-trained models using a training set of 100K samples sampled from the training set of VQAonBD 2023, with 20k samples of each question category. 

We developed multiple question-answering models using both raw HTML tables (results of table recognization in Section \ref{sec2:table_recognition}) and structured tables (results of post-structure extraction in Section \ref{sec2:structure_extraction}). In cases where the QA model returns no answer, we assign the answer as zero.

\section{Experiments} \label{sec1:experiments}

\subsection{Dataset} \label{sec2:data}
We evaluate TabIQA on VQAonDB 2023 dataset. This dataset contains document images from the FinTabNet dataset \cite{fintabnet} and relevant questions about these document images. The training and validation set's ground truth also contains table structure annotation information, i.e., bounding boxes of words, tokens, digitized text, and row and column identifiers.

Each document image may include up to 50 questions in five categories with their corresponding answers. The number of questions in each category for a single table varies based on its content and format. For Category 1, the number of questions falls within the range of 0 to 25, while for Category 2, it is between 0 and 10. Similarly, the number of questions in Category 3 ranges from 0 to 3, while for Category 4, it is between 0 and 7. Lastly, Category 5 has a range of 0 to 5 questions.

The detailed statistics of VQAonBD 2023 are described in the following sections.

\subsubsection{Document Images}
In this section, we analyze the document images of the VQAonBD 2023 dataset. The statistics of document images are reported in Table \ref{tab:doc_images}. ``Doc Images" are the number of document images, whereas the ``Blank Images" are the number of the blank page. For example, the sample ``val\_table\_image\_9684\_\_CL\_\_2014\_\_page\_54\_split\_0" in the validation set of VQAonBD 2023 is the blank page. 

The training contains 12\% blank images within the dataset, whereas validation and testing sets have blank images at less than 1\%. Despite the absence of content, related questions exist regarding these blank images. During TabIQA training phase, we exclude these samples containing blank images. In the testing phase, TabIQA returns a zero value for samples containing blank images.

\begin{table}[!ht]
\centering
\caption{Document Image Statistics on VQAonBD 2023}
\label{tab:doc_images}
\begin{tabular}{@{}lll@{}}
\toprule
           & Doc Images & Blank Images \\ \midrule
Train      & 39,999     & 5,025        \\
Validation & 4,535      & 6            \\
Test       & 4,361      & 13           \\ \bottomrule
\end{tabular}
\end{table}

\subsubsection{Questions}
The dataset used for training, validation, and testing contains 41,465, 1,254,165, and 135,825 questions, respectively. The average question length is 109.45 characters, and the average number of words in a question is 10.5. Some questions are longer than 1,500 characters. To identify named entities within the questions, we use the Spacy tool \cite{spacy}, which detects an average of 1.71 entities per question. Of these, 1.42 entities pertain to the time dimension, 0.18 are numerical values, and 0.12 have textual values.

\subsubsection{Tables}
Table \ref{tab:table_size} reports size statistics of annotated tables from the training and validation set of VQAonBD 2023. The training set comprises tables with  numbers of rows larger than those in the validation set, and the cells in the training set are also longer than those in the validation set.
\begin{table}[!ht]
\centering
\caption{Annotated Table Statistics on VQAonBD 2023}
\label{tab:table_size}
\begin{tabular}{@{}lrr@{}}
\toprule
          & Train (avg.)      & Val (avg.)         \\ \midrule
Row       & 2-77 (13.27)      & 2 -58 (12.15)      \\
Column       & 2-16 (4.57)       & 2-13 (4.44)        \\
Cell length & 3.2-161.08 (11.3) & 4.56-40.54 (11.17) \\ \bottomrule
\end{tabular}
\end{table}

\subsection{Experiment Setup} \label{sec2:setup}
\subsubsection{Baselines}
To evaluate our results, we compared them against other baselines, including TAPAS \cite{tapas}, TAPEX \cite{tapex}, OmniTab \cite{omnitab}, and the Zero model. The Zero model always returns zero for any question. 
\subsubsection{Metrics}
The VQAonBD 2023 performance metric is determined depending on answer types. If the answer types are textual values, then the Averaged Normalised Levenshtein Similarity (ANLS) is used as the metric in DocVQA \cite{DocVQA}. ANLS is designed to respond softly to answer mismatches that may arise due to OCR imperfections. On the other hand, if the answer types are numerical values, then the metric is calculated by taking a scaled Euclidean norm of the ANLS score and the percentage of the absolute difference between the predicted answer and the ground truth answer.

\subsection{Results and Discussions} \label{sec2:results}
Table \ref{tab:result} compares TabIQA's question-answering performance against other baseline models using the VQAonBD 2023 dataset. We use the fine-tuned hugging face models\footnote{Huggingface models: \url{https://huggingface.co/models?pipeline_tag=table-question-answering}} of TAPAS \cite{tapas}, TAPEX \cite{tapex}, and OmniTab \cite{omnitab} on the WikiTableQuestion dataset \cite{wtd_dataset}. In addition, we included the Zero setting, a model that always returns zero for any question. TabIQA1 represents the setting where the question-answering model is fine-tuned directly on raw HTML tables. On the other hand, TabIQA2 refers to the setting where the QA model is fine-tuned on structured tables.

\begin{table}[!ht]
\centering
\caption{Question answering performance on the validate split VQAonBD 2023 dataset.}
\label{tab:result}
\begin{tabular}{@{}lr@{}}
\toprule
Model   & VQAonBD 2023 Score  \\ \midrule
TAPAS \cite{tapas}   & 0.4138 \\
TAPEX \cite{tapex}   & 0.4390  \\
OmniTab \cite{omnitab} & 0.4421 \\ \midrule
Zero    & 0.2616 \\
TabIQA1 & 0.8808 \\
TabIQA2 & \textbf{0.8997} \\ \bottomrule
\end{tabular}
\end{table}

Regarding question-answering performance, the scores indicate that TabIQA1 and TabIQA2 significantly outperform the other baseline models, i.e., TAPAS, TAPEX, OmniTab, and the Zero model. These results suggest that TabIQA's fine-tuning the question-answering model on raw HTML tables or structured tables can significantly improve question-answering performance compared to other baseline models. It also indicates that TabIQA2 outperforms TabIQA1, which suggests that fine-tuning the QA model on structured tables can lead to better performance than fine-tuning on raw HTML tables. Overall, these results demonstrate the effectiveness of TabIQA in achieving high accuracy in answering table-related questions.

\section{Conclusion and Future Work} \label{sec1:conclusion}
This paper aims to present a new pipeline called TabIQA to answer questions related to business document images. TabIQA employs cutting-edge deep learning methods in two stages. Firstly, it extracts both the content and structural information from images of tables. Secondly, it utilizes these features to answer questions about numerical data, text-based information, and complex queries from structured tables. Experimental results on VQAonBD 2023 dataset demonstrate that TabIQA can achieve promising performance in answering questions about tables.

We plan to extend the TabIQA pipeline to handle more complex queries that require reasoning over multiple tables or information from the document's non-tabular parts. Another area for future work is to investigate the generalization capabilities of TabIQA to handle tables from different domains or document layouts. These are all potential avenues for future research that could enhance the capabilities and performance of TabIQA in real-world scenarios.
 
\section*{Acknowledgements}
The research was supported by the Cross-ministerial Strategic Innovation Promotion Program (SIP) Second Phase, “Big-data and AI-enabled Cyberspace Technologies” by the New Energy and Industrial Technology Development Organization (NEDO).

\bibliographystyle{IEEEtran}
\bibliography{IEEEabrv,ref.bib}

% \bibitem{b1} G. Eason, B. Noble, and I. N. Sneddon, ``On certain integrals of Lipschitz-Hankel type involving products of Bessel functions,'' Phil. Trans. Roy. Soc. London, vol. A247, pp. 529--551, April 1955.
% \bibitem{b2} J. Clerk Maxwell, A Treatise on Electricity and Magnetism, 3rd ed., vol. 2. Oxford: Clarendon, 1892, pp.68--73.
% \bibitem{b3} I. S. Jacobs and C. P. Bean, ``Fine particles, thin films and exchange anisotropy,'' in Magnetism, vol. III, G. T. Rado and H. Suhl, Eds. New York: Academic, 1963, pp. 271--350.
% \bibitem{b4} K. Elissa, ``Title of paper if known,'' unpublished.
% \bibitem{b5} R. Nicole, ``Title of paper with only first word capitalized,'' J. Name Stand. Abbrev., in press.
% \bibitem{b6} Y. Yorozu, M. Hirano, K. Oka, and Y. Tagawa, ``Electron spectroscopy studies on magneto-optical media and plastic substrate interface,'' IEEE Transl. J. Magn. Japan, vol. 2, pp. 740--741, August 1987 [Digests 9th Annual Conf. Magnetics Japan, p. 301, 1982].
% \bibitem{b7} M. Young, The Technical Writer's Handbook. Mill Valley, CA: University Science, 1989.
% \end{thebibliography}
% \vspace{12pt}
% \color{red}
% IEEE conference templates contain guidance text for composing and formatting conference papers. Please ensure that all template text is removed from your conference paper prior to submission to the conference. Failure to remove the template text from your paper may result in your paper not being published.

\end{document}